# Towards Improving Workers' Safety and Progress Monitoring of Construction Sites Through Construction Site Understanding⋆

Mahdi Bonyani[a,∗], Maryam Soleymani[b]

[a]*Department of Computer Engineering, University of Tabriz, Iran*
[b]*Department of Construction and Project Management – Art University of Tehran, Tehran, Iran*

**ABSTRACT**

An important component of computer vision research is object detection. In recent years, there has been tremendous progress in the study of construction site images. However, there are obvious problems in construction object detection, including complex backgrounds, varying sized objects, and poor imaging quality. In the state-of-the-art approaches, elaborate attention mechanisms are developed to handle space-time features, but rarely address the importance of channel-wise feature adjustments. We propose a lightweight Optimized Positioning (OP) module to improve channel relation based on global feature affinity association, which can be used to determine the Optimized weights adaptively for each channel. OP first computes the intermediate optimized position by comparing each channel with the remaining channels for a given set of feature maps. A weighted aggregation of all the channels will then be used to represent each channel. The OP-Net module is a general deep neural network module that can be plugged into any deep neural network. Algorithms that utilize deep learning have demonstrated their ability to identify a wide range of objects from images nearly in real time. Machine intelligence can potentially benefit the construction industry by automatically analyzing productivity and monitoring safety using algorithms that are linked to construction images. The benefits of on-site automatic monitoring are immense when it comes to hazard prevention. Construction monitoring tasks can also be automated once construction objects have been correctly recognized. Object detection task in construction site images is experimented with extensively to demonstrate its efficacy and effectiveness. A benchmark test using SODA demonstrated that our OP-Net was capable of achieving new state-of-the-art performance in accuracy while maintaining a reasonable computational overhead.

## 1. Introduction

Construction and civil infrastructure projects often involve people and equipment interacting in unique and active methods [1]. Reporting accurate progress to stakeholders may keep them informed about the situation of a project [2]. It can actually help them prevent construction delays and extra costs by controlling slippage activities. However, traditional processes of progress reporting are typically time-consuming, error-prone, slow as well as unnecessary, thus avoiding stakeholders from effectively making informed decisions [3]. In addition, traditional construction progress monitoring is a labor-intensive procedure that relies on manual data collection, documentation, and reporting the progress of a project on an ongoing basis [4]. These issues are being addressed by emerging disruptive technologies in construction [5]. Modern site operations are crucial to complete project goals and thus optimizing performance for the industrial 4.0 era [6]. Cyber-physical systems have redefined construction development into self-aware and context-aware systems based on sensing and vision systems [7]. Employing digital tools, tracking the real-time status of site operations will be possible. In fact, digitalize project progress monitoring can be achieved by automatically capturing and reporting site data [8]. Additionally, a variety of equipment poses undoubted risks to productivity, safety, sustainability, and quality of operations near overhead powerlines, moving traffic as well as trenches [9]. Monitoring equipment proficiency in the modern era must be intelligent and take into account machine anatomy and the environments in which it operates [10].

On the other hand, construction site safety has become more crucial than ever as the industry revitalizes and more infrastructure needs to be built. Moreover, it is possible to prevent a lot of accidents by wearing personal protective equipment [11]. For instance, the helmet is one of the most important pieces of personal protective equipment for keeping workers safe from falling objects [12]. Therefore, wearing helmet is legally required on many construction sites around the world [13]. Manually monitoring this makes the examiner tired and prone to misjudgment because they stare at the screen for long periods of time. New technologies and sensors are making it possible to detect helmets on construction sites by using image analysis techniques. However, challenging site conditions, such as dusty, occluding, and poor lighting, limit the capacity of markers to be detected for accurate pose estimation of articulated machines [14]. A variety of computer vision techniques are being applied to numerous fields such as automated fire detection and notification [15], construction worker safety inspections [16]. [16], Analysis analysis of ground equipment activity [17], detection of wearing the safety helmet [18], noncertified work detection [19], and identification of construction activities [20]. Digital images on construction sites are vital to the accurate detection and identification of workers and equipment [20, 16]. In one of the previous studies, detec-tors were largely based on hand-crafted features and Sliding Windows Traversing on the entire image; the algorithms also ran slowly, had insufficient accuracy, and did were not generalizable [16].

∗Corresponding author
mhd.bonyani@gmail.com (M. Bonyani)
ORCID(s): 0000-0003-0922-9656 (M. Bonyani); 0000-0003-3796-3137 (M. Soleymani)

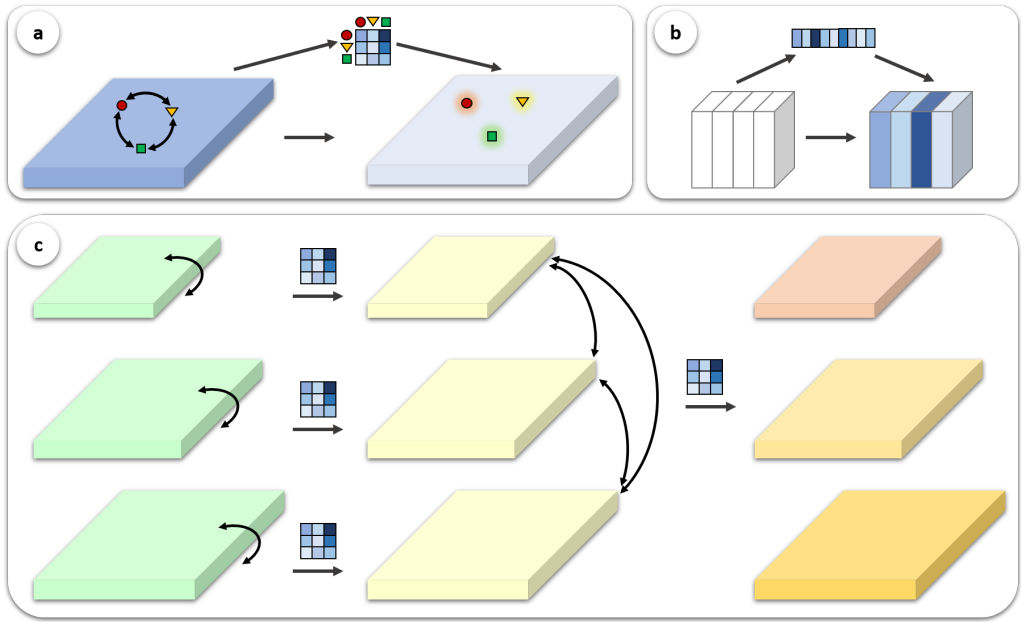

**Figure 1:** Schematic of the optimized-position methods. (a) Spatial attention mechanism. (b) Channel attention mechanism. (c) Our proposed optimized-position (OP).

Recently, deep convolutional neural networks (CNNs) have shown great promise in image processing, which has led to significant advances in object detection on construction sites. Typically, state-of-the-art approaches use either a one-stage detector (e.g. YOLO [21] or RetinaNet [22]) or two-stage detectors (e.g. R-CNNs [23, 24]). On construction sites, objects tend to be smaller, denser, and in a more complex background, and have poorer imaging quality when compared to natural scenes. As a result, using the existing natural-scene object detectors on construction sites is not an easy task with attaining an acceptable recognition performance. Accordingly, state-of-the-art methods aim to create efficient head networks [25], labeling strategies [26, 27], and adjustable dense anchor generators [28]. Furthermore, effective techniques for learning features are crucial. The performance of a model can be improved by using such methods because they provide generalized features. In order to improve the rough representation of features in CNNs [29, 30, 31, 32, 33, 34], numerous multi-scale feature adjustment methods have been proposed. Based on their conceptual structure, these attention methods can be divided into two branches: Spatial Attention (SA) methods and Channel Attention (CA) methods. The first category (such as SA modules used in [29, 35, 31], recurrent attention structures [32], self-awareness mechanisms [36], and non-local operation [37]) can be mapped to a global context by calculating the similarity between each specific feature position and the remaining features [38, 39] as shown in Fig. 1. A long-range dependence information can be extracted from each pixel by such an operation. It is illustrated in Fig. 1 that each channel can capture a weight related to its significance in object detection (e.g., CA module , Squeeze and Excite block, [29, 40]). This weight can then be integrated into the model by channel reweighting.

We argue that most existing feature adjustment methods for channel adjustment are not sufficient despite the success of attention-based object detection methods. Therefore, they cannot use channel relation to capture channel feature map dependencies, which have been empirically demonstrated to be helpful in various computer vision recognition problems. Despite existing methods to differentiate channel weights based on channel attention, modules such as global average-pooling and max-pooling based on their channel feature maps cannot assure adequate relations between channels.

This article proposes a simple but effective optimized-position (OP) attention for enhancing channel relation through feature transformers that can adjust the verification weights for each feature channel related to global feature affinity-pairs, in order to address these problems, such as different objects with various sizes and complex and various backgrounds in construction site and limitations of existing attention methods in adjusting features. According to Fig 1(c), OP is an active feature relation mechanism that introduces channel-based feature dependencies. It considers the pyramid layers as the "channel" of the overall pyramid features, including the inner and interlayers of the pyramid . OP involves two steps: first, dot-product operations are used to compute feature similarity between each channel and the rest of the channels. Following that, the adjustment is used to aggregate the channel weights together to represent each channel. It contains richer information about long-range channel dependency than the input feature maps, but it has the same spatial size as the input feature maps. As a solution to typical construction site imaging problems, we propose rearranging pyramid OP within and between layers to adjust local and global features within the images.

Any deep neural network can connect to the OP attention. OP-Net refers to a CNN model deployed with the proposed OP module. In Fig. 2, you can see the overall architecture. Extensive experiments are conducted on object detection tasks to demonstrate its effectiveness and efficiency. As a result of experimental results comparing object detection on construction sites (SODA) [41], our proposed OP-Net is able to significantly improve object detection compared to baseline methods and achieves state-of-the-art performance in accuracy, while consuming relatively little computational resources. OP-Net also achieves state-of-the-art performance in experiments on SODA [41]. In summary, the contributions of the proposed method are as follows:

- We propose an end-to-end network for object detection that utilizes the different proposed attention module collaborative learning approach to capture a more comprehensive feature.

- A novel and effective approach to OP attention is proposed that involves implementing within and between feature pyramid layers to enhance pyramid representation, and involves enhancing channel relations in the form of feature transformers.

- We propose an OP-Net, which can achieve state-of-the-art object detection performance on challenging benchmark for SODA as construction site images.

## 2. Related works

AI object recognition, which allows the identification of multiple objects, has recently emerged in comparison with semantic segmentation and classification, which can identify only a single object. However, it is impossible for object recognition to pinpoint the location of an object despite being able to distinguish one from another. Multiple objects and their locations can be detected with object detection. According to Son et al. [42], Faster-CNN with bounding boxes was used to detect the presence of workers in the construction industry. It is still feasible to identify hazard types with this classification approach though it can only detect one object at a time, which limits its direct practical application for hazard identification. There is also no standard benchmark for the existence of hazards because their study did not indicate their probability level. The construction workers' activities were detected by CNN by Luo et al. [20]. However, the safety risks associated with construction were not investigated. Regarding many construction site safety risks, it is necessary to develop object detection based on real-life situations. As-built models are overlayed over as-planned BIM models and then compared to determine the volume of work completed. The output of this process is then converted into percent complete. In other words, identifying specific construction features using the targeted BIM model for a specific type of building structure, e.g. an RCC structure. From a vision dataset, various construction features are identified and measured according to their quantity or relative position. Object detection is also a method that provides a certain level of automation depending on the type of technique that is being used, as well as BIM registration.

### 2.1. Object detection for progress estimation

In order to estimate progress, the built environment must be recognized for various features or objects. By superimposing both models, it is not possible to find useful information about the differences between the as-built and as-planned datasets. To retrieve information, as-built point clouds must be recognized, identified, or classified so that information from as-planned and as-built BIMs can be compared. There have been a number of methods proposed to classify, identify, and detect the various construction components, such as walls, panels, tiles, ducts, doors, windows, and furniture [43, 44, 45]. This includes Mask R-CNN, Voxel, Point Net, DeepSORT, Probabilistic Model, OpenGL, Surface-based recognition, timestamps, point calculations, segmentation by color thresholding, and color images. Object-based models for estimating progress can provide useful information with these techniques. Besides comparing as-built with as-planned, some studies have also compared equipment cycles [46], material classification [46], material usage [47, 48], and installation speed [49].

### 2.2. Object detection for risk and safety estimation

Fang et al. [18] used more than 100,000 construction worker images, who are pioneered a faster R-CNN approach for detecting workers without helmets. The newly developed algorithm allowed Custom Vision to train its system based on only 50 images, saving time and money by automating the generation of a database of safety risk images for training, which allows the system to identify many more hazards. A 99.8% accuracy rate was achieved by Li [50] using 25 high-quality photos to train a system to recognize electrical wires and holes on sites. A custom vision object detection algorithm, however, has not been used for detecting many different types of safety risks on construction sites. An integrated holographic hybrid reality has not been developed for safety inspections and training or a neuroscience approach has not been applied to construction risk decision-making.

## 3. Methodology

### 3.1. Adjustable Channel Attention

Channel attention (CA) module weighs the similarity matrix to emphasize the most important channels by leveraging the interdependencies between them. In CA, queries (Q) are triggered by keys (K), and values (V) are triggered by the queries within each set of characteristic maps M. $\mathbf{M}'$ and $\mathbf{M}$ have similar scales based on the original $\mathbf{M}$. CA can be implemented by specifying W, H, and C for each set of feature maps $\mathbf{M} \in \mathbb{R}^{W \times H \times C}$. This can effectively be achieved by building a feature map using a set of features. CA implementation can be formulated as:

$$\text{sim}_{i,j} = G_{\text{sim}}\left(\mathbf{q}_i, \mathbf{k}_j\right)$$
$$\mathbf{w}_{i,j} = G_{\text{nom}}\left(\text{sim}_{i,j}\right)$$
$$\mathbf{M}'_i = \sum_j G_{\text{mul}}\left(\mathbf{w}_{i,j}, \mathbf{v}_j\right)$$

Where $\mathbf{q}_i, \mathbf{k}_j, \mathbf{v}_j$ represent input of CA and $\text{sim}_{i,j}, \mathbf{w}_{i,j}$, and $\mathbf{M}'_i$ denote similarity, weight and output, respectively. In this example, $\mathbf{q}_i = g_q\left(\mathbf{M}_i\right) \in \mathbf{Q}$ denotes the i-th query, $\mathbf{k}_j = g_k\left(\mathbf{M}_j\right) \in \mathbf{K}$ denotes the second key/value pair, and $\mathbf{v}_j = g_v\left(\mathbf{M}_j\right) \in \mathbf{V}$ denotes the third key/value pair. $g_q(\cdot), g_k(\cdot)$, and $g_v(\cdot)$ denote the queries/keys/values channel transformation functions [36, 51]. There are two channel features in $\mathbf{M}$ : $\mathbf{M}_i$ and $\mathbf{M}_j$; $G_{\text{sim}}$ represent the dot product similarity functions; KK and $G_{\text{mul}}$ denote matrix dot multiplication; $\mathbf{M}'_i$ represents the ith channel feature in $\mathbf{M}'$, and the response of $\mathbf{M}'_i$ is computed by jth ones that enumerate all channels possible. Despite the ability of CA to generate different weights for different channels, coarse operations (i.e., without grouped feature representations [52, 51, 53, 54]) cannot allow all channels to communicate sufficiently, which has been empirically demonstrated to be crucial to a variety of computer vision tasks. Due to this limitation, features cannot be represented well.

### 3.2. Optimized-Position (OP)

Based on global feature affinity-pairs, OP can be used to enhance feature channel relation by setting optimized weights adaptively based on the feature channel affinity pairs. Fig. 2 shows its detailed structure. By combining the multi-head representations and concatenating them with the optimized features, we produce enhanced feature maps by using a convolution layer based on the transformer mechanism. To enable richer channel feature representations, we deploy the multi-head architecture. Multi-head can provide more feature selection when extracting features in ViT [55] and DETR [36]. Using more than one head to complement features is more efficient than learning the same contents in one head. Based on their analysis work [56], important multi-head models have one or more specialized functions that are interpretable, demonstrating the need for multi-head models.

Firstly, we divide the channel dimension into P parts. Each structure in each head is an OP module ( B is the batch size), based on the divided features with shape $(B, C/P, H, W)$. A similarity matrix in the form of $(B, C/P, C/P)$ exists for the n-th head module, which is expressed as follows:

$$\text{sim}^n = \begin{bmatrix} w^{pC/P, pC/P} & \cdots & w^{(p+1)C/P, 0} \\ \vdots & \ddots & \vdots \\ w^{0,(p+1)C/P} & \cdots & w^{(p+1)C/P,(p+1)C/P} \end{bmatrix}$$

In this case, every w represents a scalar of similarity that can be learned. Following the concatenation of the partial results from these head modules, we obtain the holistic output feature maps from the original feature maps with the same shape. A process similar to the one mentioned above can be described as:

$$\text{Weight} : \mathbf{w}^n_{i,j} = G_{\text{nom}}\left(\text{sim}^n_{i,j}\right)$$
$$\text{Partial result} : \mathbf{M}^n_i = \sum_j G_{\text{mul}}\left(\mathbf{w}^n_{i,j}, \mathbf{v}_{j,n}\right)$$
$$\text{Holistic output:} \mathbf{M}' = G_{\text{con}}\left(\mathbf{M}^n_i\right)$$

Here, the weights of each channel feature and its normalized version are denoted by $\text{sim}^n_{i,j}$ and $\mathbf{w}^n_{i,j}$. Several channel features contribute to the calculation of the ith channel feature. An nth head's j-th value is denoted by $\mathbf{v}_{j,n}$. Concatenating features in the channel dimension is done with $G_{\text{con}}$. A multi-head OP with $O\left(PC^2\right)$ computational complexity has lower computational complexity than the previous transformer-based approaches, which have $O\left(PH^2W^2\right)$ computational complexity. We propose OP implementations based on pyramid features because they offer three advantages over CA. Communication within and between feature pyramid layers is promoted by OP, whereas most previous methods capture long-range dependencies between features within and across space. In OP, features are represented in different feature spaces as a result of the multi-head structure [57, 51]. Consequently, OP can enhance the representation of features. A construction image is analyzed by OP to detect objects. In Section IV-B, OP proposes more accurate head network proposals by increasing feature pyramid representation in construction images to solve complex background and poor imaging problems. In both oriented and horizontal tasks, OP improves state-of-the-art performance dramatically (see Section IV-C). As follows are two OP implementations of a pyramid OP show with a base OP implemented on top.

### 3.3. Base OP

We can extract feature maps from arbitrary construction images using a fully convolutional network. These feature maps can be used by OP directly to adjust weights for each channel and improve communication channels. Fig. 2(b) depicts the detailed architecture of each level of the feature pyramid (i.e., feature maps with the same scale). Base OP is implemented on the basic feature maps since it is based on the basic feature maps. A base OP works on a backbone network and is a general unit. A wide range of downstream recognition tasks can be supported by this method, while other existing head-network-based methods [58, 33] are more task-specific. Section IV-B shows the ablation experiments that demonstrate how our base OP improves feature extraction.

### 3.4. Modified Pyramid OP

There has been extensive research on the effectiveness of feature pyramids in the computer vision field [59, 22, 60]. A rearranged pyramid OP (MP-OP) is proposed here, which shows how to implement our OP on a pyramid feature [61, 62, 63]. We have developed an efficient, low-complexity, and more parameter-light MP-OP, which is applied to the in-network feature pyramid (see Section IV-A for details). A table of MP-OP modules is shown in Fig. 2(c). These modules extract feature pyramids from the feature pyramid network [59]. A level in the feature pyramid is viewed as a small piece of the input image's features, i.e., only a fraction of the input image's features is captured at each level. The combination of global and local information is crucial in feature extraction in order to highlight the most suitable feature in the channel dimension. Based on works [59, 60], MP-OP is used to weight different features across pyramid levels $\mathbf{M}_{S2-S6}$. The feature pyramid is illustrated in Fig. 2(c) with OP applied between the five levels in order to fully convey the information of each level. We begin by reducing the channel dimension and activating interpolation on pyramid features $\mathbf{M}_{S2-S6}$ to produce features of the same scale (same scale as S2), and then we concatenate them into $\overline{\mathbf{M}}_{S2-S6}$, which is expressed as:

$$\overline{\mathbf{M}}_{S2-S6} = G_{\text{intp}}\left(\mathbf{M}_{S2-S6}\right)$$

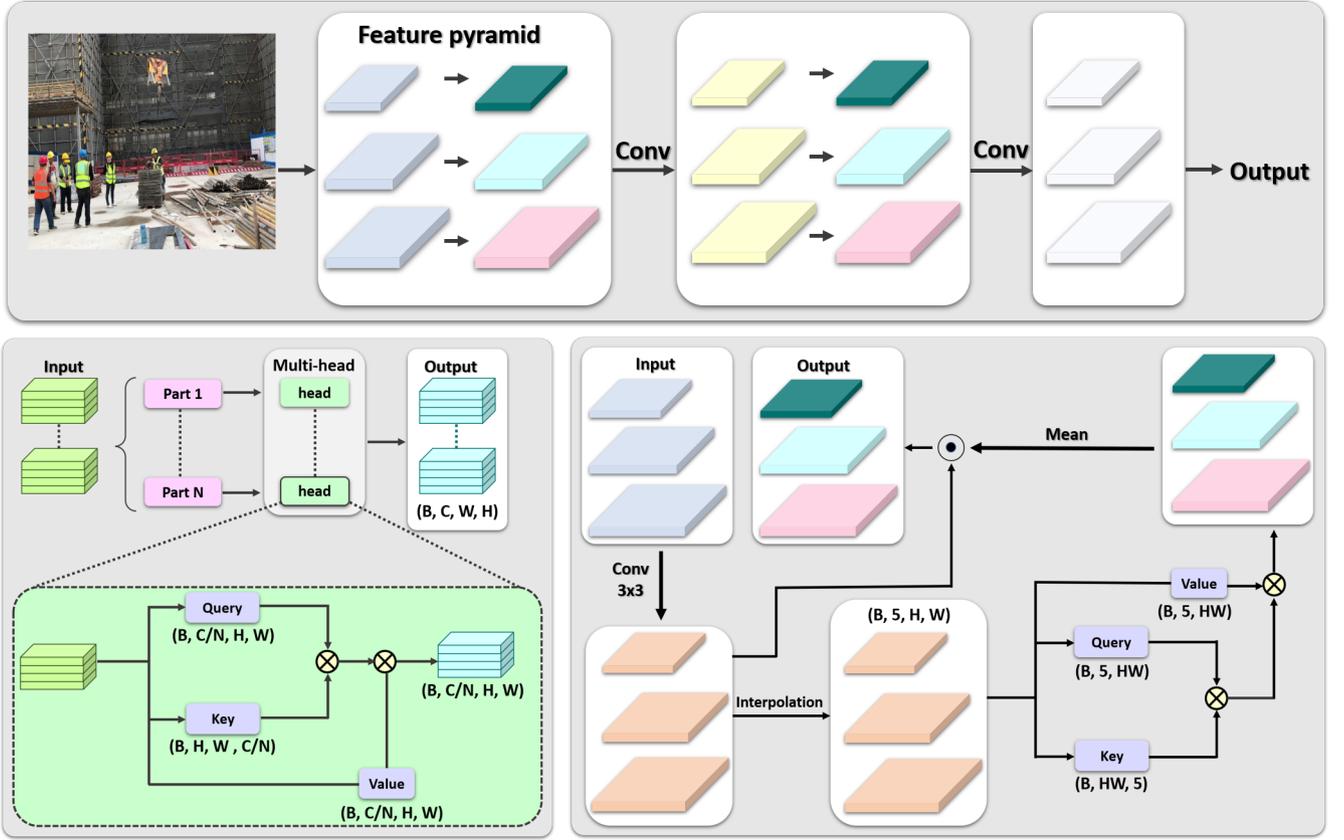

**Figure 2:** Our proposed OP network (OP-Net), which deploys OP both on intralayer feature maps and on feature pyramids.

An interpolation function $G_{\text{intp}}$ is used to reduce channel dimensions and scale. Output feature $\overline{\mathbf{M}}_{S2-S6}$ has a shape of $(B, 5, H_{s2}, W_{s2})$. After that, MP-OP produces $\overline{\mathbf{M}}'_i$ by learning the weight between the query and the key from input $\mathbf{q}_i, \mathbf{k}_j$, and $\mathbf{v}_j$. Interactions can be expressed as

$$\text{Input}: \mathbf{M}_{S2-S6}$$
$$\text{Interpolation}: \overline{\mathbf{M}}_{S2-S6}$$
$$\text{Extraction}: \mathbf{q}_i, \mathbf{k}_j, \mathbf{v}_j$$
$$\text{Similarity}: \text{sim}_{i,j} = G_{\text{sim}}\left(\mathbf{q}_i, \mathbf{k}_j\right)$$
$$\text{Weight}: \mathbf{w}_{i,j} = G_{\text{nom}}\left(\text{sim}_{i,j}\right)$$
$$\text{Output}: \overline{\mathbf{M}}'_i = \sum_j G_{\text{mul}}\left(\mathbf{w}_{i,j}, \mathbf{v}_j\right)$$
$$\text{Holistic utput}: \overline{\mathbf{M}}^{\text{rpcg}}_{S2-S6} = G_{\text{con}}\left(\overline{\mathbf{M}}'_i\right)$$

A feature $\overline{\mathbf{M}}'_i$ in $\overline{\mathbf{M}}^{\text{rpcg}}_{S2-S6}$ is at the ith level, and $(B, 5, H_{s2}, W_{s2})$ is the shape of $\overline{\mathbf{M}}^{\text{rpcg}}_{S2-S6}$. Pyramid features are fully realized $\overline{\mathbf{M}}^{\text{rpcg}}_{S2-S6}$, but we need to find a way to feedback to them.

Furthermore, a number of methods have been used in visual recognition to verify the effectiveness of local and global information combined, and our method is a global approach. As a result, we chose to combine our MP-OP method with the existing local channel attention method. We choose classic channel attention [64] for this study. Hence, our pro-posed MP-OP module has the structure as:

$$\text{Weight}: \overline{\mathbf{M}}_{S2-S6}^{\text{avg (rpcg)}} = G_{\text{avg}}\left(\overline{\mathbf{M}}_{S2-S6}^{\text{rpcg}}\right)$$
$$\text{Scale}: \overline{\mathbf{M}}'_{S2-S6} = \overline{\mathbf{M}}_{S2-S6}^{\text{avg (rpcg)}} \otimes \mathbf{M}_{S2-S6}$$
$$\text{Output}: \overline{\mathbf{M}}_{S2-S6}^{\text{out}} = G_{\text{conv}}\left(\overline{\mathbf{M}}'_{S2-S6} \oplus \mathbf{M}_{S2-S6}\right).$$

S2 through S6 are the five outputs from $\overline{\mathbf{M}}_{S2-S6}$. A weighted average of $\overline{\mathbf{M}}_{S2-S6}$ is called $\overline{\mathbf{M}}_{S2-S6}^{\text{rpcg}}$. For each pyramid's level, the mean value is used as the weighting parameter, which is then resized to the same scale as the original level feature used by $G_{(\text{avg})}$ to distinguish between scales. A matrix cross multiplication is performed by $\otimes$, and a channel concatenation is performed by $\oplus$. As with the original feature pyramid, $\overline{\mathbf{M}}'_{S2-S6}$ is adjusted to the same size. In order to restore the original size of the channel, we get the output $\overline{\mathbf{M}}_{S2-S6}^{\text{out}}$ from convolution $G_{\text{conv}}$.

### 3.5. Network Architecture

For construction image object detection tasks, OP can boost the model's ability to learn richer communication information among feature channels. The purpose of this article is to develop an OP-Net able to detect oriented and horizontal objects in construction images. In Fig. 2, you can see the overall architecture. We propose OP-Net as a method for transforming pyramid features based on OP and MP-OP as shown in Fig. 2. A ResNet [65]-based backbone is deployed on top of [25], which has been pretrained on ImageNet [66]. Our feature pyramid is then produced based on the feature pyramid network [59]. Our feature pyramid begins by applying base OP to each of the feature maps. A new feature pyramid is then created in which local and global communication is realized. A 3x 3 convolution reduces the dimension of the concatenated features maps to 256 channels by concatenating the original feature maps and adjusted ones together. To detect horizontal objects, we use a standard faster R-CNN [24], which is derived from the head network of the RoI transformer [25].

### 3.6. Experiments

Our proposed method is shown to be effective and efficient both on oriented object detection tasks in construction images and horizontal object detection tasks. Firstly, in Section IV-A, we examine the experiments settings, including datasets, image size, baseline model, hyperparameters, implementation details, and evaluation metrics. We then show some experimental results in Section IV-B, including both quantitative and qualitative results, followed by a comparison of results with state-of-the-art methods in Section IV-C.

### 3.7. Train Setup

As part of our experiments, we select a challenging dataset that is a large-scale dataset from our the SODA dataset [41]. Object detection can be done both horizontally and orientedly with SODA.

It includes oriented and horizontal bounding boxes and is one of the largest construction image datasets for object detection. 188,282 objects are annotated in 2806 construction site images, which were captured by different platforms and sensors, and 15 object categories are common across all the images in the SODA database like Slogan (SL), Fence (FE), Hook (HK), Hopper (HO), Electric Box (EB), Cutter (CU), Handcart (HA), Scaffold (SC), Brick (BR), Rebar (RE), Wood (WO), Board (BO), Helmet (HE), Vest (VE), and Person (PE). The images are larger than 1920×1080 pixels and contain a wide variety of objects oriented differently and captured in different scales. Our classification consists of randomly dividing the original images into three sets: training, testing, and validation, according to the method described in [41].

**Table 1**
A comparison of the effectiveness of the proposed methods with different networks for extracting features on SODA [41].

| Backbone | +Ours | GFLOPs / FPS | #Params (M) | mAP(%) |
|---|---|---|---|---|
| ResNet-50 | ✗ | 212.30/25.1 | 41.20 | 79.91 |
|  | ✓ | 365.30/14.7 | 42.98 | 83.86 |
| ResNet-101 | ✗ | 290.31/21.0 | 60.21 | 80.57 |
|  | ✗ | **443.31/13.3** | **61.98** | **85.27** |
| ResNet-152 | ✗ | 366.33/17.2 | 75.88 | 80.74 |
|  | ✓ | 523.19/12.3 | 77.65 | 84.31 |

*Note*: "+ Ours" denotes the architecture of our proposed base OP and MP-OP attention on the backbone networks.

**Table 2**
A comparison of the effectiveness of our proposed multi-head methods on SODA [41].

| Baseline | OP | Multi-head | GFLOPs / FPS | #Params (M) | mAP(%) |
|---|---|---|---|---|---|
| ✓ | ✓ | ✗ | 461.49/11.9 | 62.34 | 84.62 |
| ✓ | ✓ | ✓ | **443.31/13.3** | **61.98** | **85.27** |

*Note*: "OP" denotes our proposed OP attention on the backbone. "Multi-head" indicates combine multi-head structure in OP blocks.

In this study, we are using Faster R-CNN [24], which is combined with ResNet-101 as our backbone. A feature pyramid with predefined anchors for pyramid level S2-S6 is constructed using the FPN [59] as neck network. A rotated head network, RoI-transformer [25], is used in oriented object detection to convert horizontal proposals into rotated ones. As outlined in [25, 41], all experimental settings and parameters are strictly consistent. End-to-end training is applied to the entire network. It is necessary for the fairness of comparisons to adjust hyperparameters even though it is conducive to further improving model performance. We set anchor size as follows in [25] and [28], for SODA, with aspect ratios of [1/2,1,2] and anchor strides of [4,8,16,32,64] at each pyramid scale. Our ablation studies are based on SODA, which does not include any data augmentation. This allows for a fair comparison and trial of the proposed method. We only add random rotation to augmentations like [25], [28], and [33] compared to SOTA methods on SODA . The number of multi-heads in the base OP can be determined by P, which is a hyperparameter for multi-head. If P is high, the dividing feature reduces the ability of the channel to relationship with each other. We set S to two in the final network based on the parameter settings of previous work [60].

**Table 3**
A comparison of different attentions on the SODA [41] test set and ResNet-101 [65] is the backbone.

| Base | ✓ | ✓ | ✓ | ✓ |
|---|---|---|---|---|
| Base OP | ✗ | ✓ | ✗ | ✓ |
| OP-MP | ✗ | ✗ | ✓ | ✓ |
| SL | 70.36 | 72.81 | 74 | 76.23 |
| FE | 72.61 | 70.14 | 74.67 | 77.54 |
| HK | 91.01 | 92.11 | 90.3 | 94.73 |
| HO | 90.1 | 91.77 | 93.49 | 95.64 |
| EV | 77.44 | 75.82 | 78.87 | 80.31 |
| CU | 87.29 | 88.14 | 90.35 | 91.64 |
| HA | 90.06 | 91.8 | 90.12 | 93.48 |
| SC | 89.32 | 91.12 | 92.37 | 93.16 |
| BR | 88.71 | 89.97 | 90.57 | 92.95 |
| RE | 67.39 | 69.13 | 73.12 | 73.7 |
| WO | 89.59 | 90.71 | 92.68 | 94.27 |
| BO | 86.37 | 90.05 | 91.19 | 93.5 |
| HE | 62.18 | 60.86 | 64.8 | 66.34 |
| VE | 73.87 | 75.02 | 77.5 | 78.12 |
| PE | 72.19 | 72.98 | 72.81 | 77.51 |
| mAP | 80.57 | 81.5 | 83.12 | 85.27 |
| #Params | 60.21 | 60.78 | 60.8 | 61.98 |
| GFLOPs/FPS | 289.25/20.8 | 340.8/19.1 | 341.14/17.1 | 444.22/13.4 |

As a result of our study, the learning rate is initially 0.005 and the SGD optimizer performs a weight decay of 0.0001 and momentum decay of 0.95. We set the training epoch for SODA to 80. Neither multiscale input nor TTA are used in the testing step. Furthermore, the Colab GPU is used for the experiments. According to [41], the model is evaluated and the result distribution is analyzed using the mean average precision (mAP) of each category and overall. Moreover, $GFLOPs/FPS$ and model parameters (#Params) are employed for verifying model efficiency, which is used for determining model computational complexity and runtime efficiency.

## 4. Ablation Study

Our study is based on SODA[41] and attempts to detect objects in construction site images in the following ways:

1. Testing the efficacy and efficiency of various feature extraction networks using our proposed approaches;
2. Checking the efficiency of the two proposed attention using base OP and MP-OP;
3. Comparing our proposed methods with different attention structures;
4. Improving the detection of construction objects using RPN input;
5. Showing that different scales have mismatched error rates;
6. demonstrating some visual results.

### 4.1. Different Feature Extraction Networks

In Table 1, the experimental results on SODA's test set, comprising ResNet-50, ResNet-101, and ResNet-152, show different backbone network results. Using the combination of our module and $GFLOPs/FPS$, #Params and mAP, we compare improvements in $GFLOPs/FPS$, #Params and mAP. Our attentions are combined with the backbone, so we observe a 3.95, 4.7, and 3.57 percent increase in mAP for ResNet-50, ResNet-101, and ResNet-152, respectively. Furthermore, model efficiency is compared based on #Params and $GFLOPs/FPS$. It involves around 155 GFLOPs increment, with an average of 1.80M model parameters, and a reduction in performance of around 5-10 FPS. Our experiments are based on ResNet-101 due to its mAP and computational complexity.

### 4.2. Proposed Units

According to Table 3, we have calculated the combined performance of our proposed units on ResNet-101. The bounding box mAP improves 0.93% and 2.55% when OP base and MP-OP are used. To illustrate the trend of performance change, Fig. 3 shows the mAP radar chart for each category. Our proposed OP-Net model can increase mAP by as much as 4.7% when base OP and MP-OP are combined (i.e., our proposed OP-Net), with some categories showing very significant improvements (BO 7.13%, RE 6.31%, and SL 5.87%). By using base OP, and MP-OP, we were able to further improve feature presentation capabilities. According to the model efficiency, the base OP brings 0.59M model parameters with 51.53 GFLOPs, while the MP-OP brings 0.61M with 51.89 GFLOPs. This combination produces an increment of 154.95 GFLOPs and 1.79M model parameters. As the result of our proposed OP, GFLOPs increases from 289.25 to 444.22, after calculating the similarity matrix to features. According to Table 2, when adding the multi-head structure in our OP module, #Params reduce 0.36M and mAP increments by 0.65%.

### 4.3. Improving RPN Input for Construction Object Detection

A significant benefit of OP-Net is its ability to address complex background problems and low image quality. Construction images have more complex backgrounds due to overhead shots from different angles that show geological structures, different size objects, and different object categories. It is detrimental to learning object features in construction object detection when the imaging quality is poor. This directly affects the training of the modeling algorithms. As a result, we reorganize pyramid OP by implementing base OP on pyramid features. Depending on the maximum response layer, pyramid features generate proposals smaller or larger than those in the region proposal network (RPN)[24]. Because of this, the ROI module will be more difficult to train the detection box when the object proposals are accurate or not. By using OP-Net, the model is able to learn more detailed relation information both between layers and within layers of pyramids. The OP operation should be performed for pyramid features before adding them to the region proposal network.

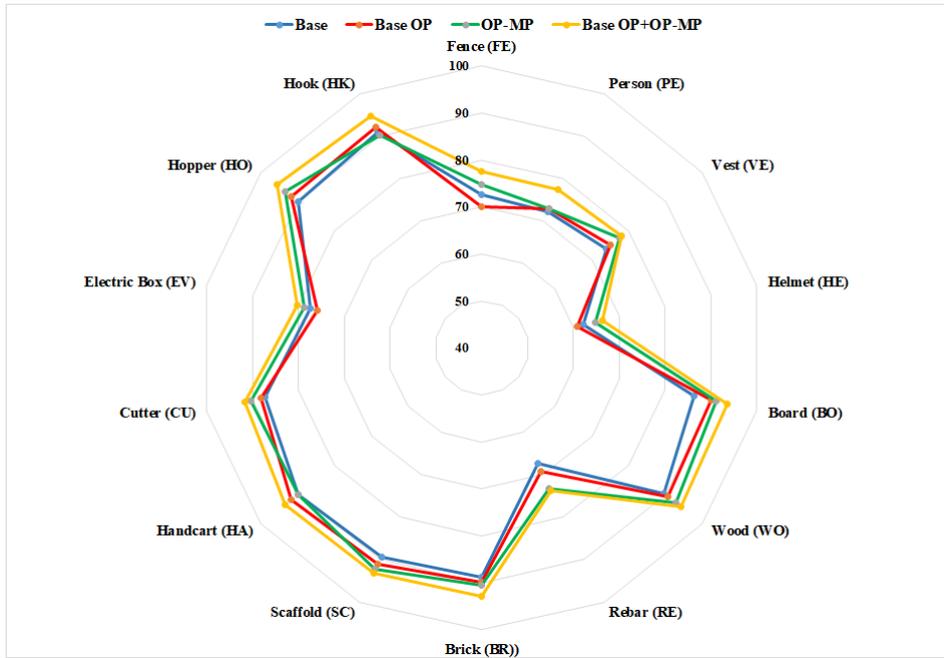

**Figure 3:** Radar chart for each category of object in SODA [41] dataset. Different detectors are represented by different colored lines. This figure represents the mAP value.

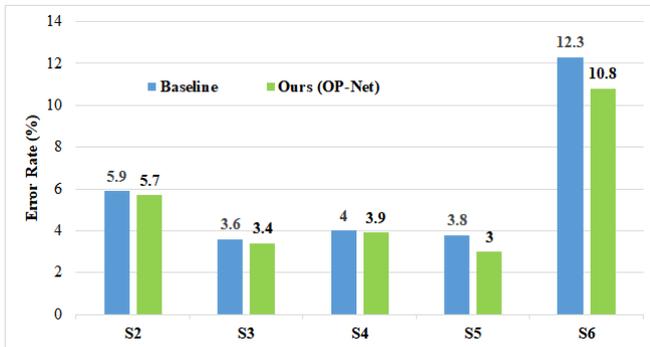

**Figure 4:** A comparison of the error rate for the baseline method of SODA [41] and the proposed method. The lower the better.

### 4.4. Error Rates

As a means of demonstrating the impact of the proposed method on each feature level, we define mismatching error rates at different scales within the feature pyramid, that is, objects chosen at different levels are not always consistent with the ground truth. According to Fig. 4, deploying our proposed method (i.e., combining base and MP-OP) resulted in a reduction of mismatching error rates for the layers in the feature pyramid. It is evident that the error rates for high-level features are lower than those for low-level features that are meant for small objects. In level S2 to S6 there is a 0.2% reduction, 0.2% reduction, 0.1% reduction, 0.2% reduction, and 1.5% reduction in error rate. As a result, our method's effectiveness can be further confirmed.

## 5. Conclusion

The detection of construction objects is complicated due to a complex background and poor image quality. Space-time feature adjustments are typically approached with elaborate attention mechanisms that are arduous in their computational complexity. For enhanced channel relation, we proposed OP attention that could determine adjust weights by channel. Through extensive experiments on construction image object detection, we implemented OP on a feature pyramid network that is the backbone of a standard object detection network. A small computational overhead is required for the proposed OP-Net to achieve state-of-the-art performance on challenging benchmarks. mAP radar charts displayed robust trends for object detection in each category. As we explore OP-Net's application to more natural scenes, we will explore applying it to different types of subjects. As well as semantic segmentation, object reidentification and other visual tasks, OP-Net is being explored in other directions.

## 6. Disclosures

The authors declare no conflict of interest.